# Semantic Web Enabled Geographic Question Answering Framework: GeoTR

Ceren Ocal Tasar[1], Murat Komesli[2,*], Murat Osman Unalir[3],

[1] Yaşar University, Department of Computer Engineering, Universite Cad. 35, 35100, Bornova, Izmir, Turkey., Turkey. E-mail: ceren.ocal@gmail.com; ORCID: 0000-0002-0652-7386

[2] Yaşar University, Department of Management Information Systems, Universite Cad. 35, 35100, Bornova, Izmir, Turkey. E-mail: murat.komesli@yasar.edu.tr; ORCID: 0000-0002-8240-5540

[3] Ege University, Department of Computer Engineering, Universite Cad. 9, 35100, Bornova, Izmir, Turkey. E-mail: murat.osman.unalir@ege.edu.tr; ORCID: 0000-0003-4531-0566

[*]Corresponding Author: Prof. Dr. Murat Komesli, Yaşar University, Department of Management Information Systems, Universite Cad. 35, 35100, Bornova, Izmir, Turkey. E-mail: murat.komesli@yasar.edu.tr; Tel (Office): +902325707817; Mobile: +905055718202

# Semantic Web Enabled Geographic Question Answering Framework: GeoTR

**Abstract**

With the considerable growth of linked data, researchers have focused on how to increase the availability of semantic web technologies to provide practical usages for real-life systems. Question answering systems are an example of real-life systems that communicate directly with end-users, understand user intention and generate answers. End users do not care about the structural query language or the vocabulary of the knowledge base where the point of a problem arises. In this study, a question-answering framework that converts Turkish natural language input into SPARQL queries in the geographical domain is proposed. Additionally, a novel Turkish ontology, which covers a $10^{th}$-grade geography lesson named "Spatial Synthesis: Turkey", has been developed to be used as a linked data provider. Moreover, a gap in the literature on Turkish question answering systems, which utilizes linked data in the geographical domain, is addressed. A hybrid system architecture that combines natural language processing techniques with linked data technologies to generate answers is also proposed. Further related research areas are suggested.

**Key Words:** question answering systems, linked data, ontology development, natural language processing

## 1. Introduction

As the importance of technology and the amount of stored data increases, fast and easy access to information for end-users has become a never-ending challenge. One of these challenges is that, as natural language is the most common way for human beings to communicate as end-users, user intention in the form of natural language must be transformed into a structural format and represented with machine-understandable notation. Tokenizing natural language input, analysis of each token, converting it into a machine-understandable representation, and generating answers or outcomes based on user intention are the major elements focused on by researchers in this area. Natural language processing (NLP), one of the research fields within artificial intelligence, can be used to solve some of these aforementioned issues.

Processes that perform the tasks of tokenizing natural language input and carrying out linguistic analysis on each token and their components are the main phases of NLP. However, despite the deep analysis outcomes of NLP, there are some weaknesses in enriching input semantically. Therefore, semantic technologies are required to analyse the outcomes of NLP and to focus on supporting techniques and methodologies to achieve an upper-level understanding of expressions (Berners-Lee et al. 2001). The technology behind semantic analysis consists of ontologies that are composed of relationships and taxonomies to represent a specific conceptualization in a specific domain. Ontologies play an essential role in sharing and exchanging knowledge between different platforms of different domains. An ontology provides flexibility and interoperability opportunities, which has resulted in a significant increase in the number of studies of the ontology concept, OWL or RDF representation, and ontology query languages like RDQL and SPARQL.

Advances in NLP techniques have also developed on account of improvements in semantic technologies. Recently, a new discussion has arisen about how to understand natural language input and generate appropriate answers with hybrid solutions, using a combination of NLP techniques and semantic technologies. Researchers have attempted to improve NLP outcomes by applying semantic enrichment. Another important discussion concerns how NLP contributes to ontology learning, ontology querying, and multilingual ontology mapping. Hybrid solutions involve a reciprocal relationship between NLP and semantic technologies (R. Q. Guo et al. 2009).

Understanding a sentence involves two main phases: morphological analysis and semantic analysis (Y. Guo et al. 2011). Morphological analysis focuses on determining the structure of words, the relationships between words, POS (part of speech) tags (noun, verb, adjective, etc.), and their positions in the sentence (subject, object, predicate, etc.). Named entity recognition (NER) is one of the NLP techniques for semantic enrichment. Entities in an expression are tagged by using predefined categories, such as person, organization, date, location, number, etc. in the NER model (Collobert et al. 2011). The generated outcome is combined with conceptualization and linked data defined in ontologies. The hybrid approach provides an opportunity to use ontologies as data sources for question answering systems that have a natural language input format. A knowledge requirement denoted in natural language is transformed into ontology query language. Without dealing with additional technical details relating to the ontology, such as query language, vocabulary or hierarchical structures, even end-users can utilize the knowledge presented in ontologies (Bernstein et al. 2005). This approach bridges the gap between end-users and semantic technologies and also provides practical implications for ontologies.

This study firstly outlines the Question Answering Framework (QAF) over linked data, presents a detailed comparison between them, and proposes a Turkish geographical question answering framework that combines NLP techniques and semantic web technologies over linked data. The current section introduces the problem, concepts, motivation, and discussions. The following section includes background information and related work. Section 3 defines the natural language-aware ontology development process. Following this, section 4 describes the proposed framework and developed geographical ontology by dividing it into 3 subsections: question pre-processing; query formulation; and experimental study. Conclusions and future work are set out in Section 5. The contributions of this study to the literature are also discussed in the final section.

## 2. Background Information and Related Work

12 question answering systems or frameworks over linked data with cutting edge approaches are examined to give an overview of the literature.

Habernal and Konopík introduced a Semantic Web Search Using Natural Language (SWSNL) system. SWSNL employs ontologies in order both to store data and domain structure, and to return answers for users' search queries in natural language. The pipeline described for the system holds the phases for pre-processing, semantic analysis, semantic interpretation, and executing SPARQL to generate answers. The authors claim there are significant differences between their approach and other similar systems in that users can formulate natural language queries in more than one sentence, and stochastic semantic analysis model pre-processing is handled without requiring any syntactic parser. SWSNL has been tested for three domains in different languages: accommodation in English, public transportation in Czech, and a subset of the English ATIS dataset. Serving different domains and languages addressed the portability feature of SWSNL in their study (Habernal and Konopik 2013).

Xser is a question answering system that generates answers over DBpedia by transforming natural language questions into linked data query representation. Unstructured natural language input is converted into structured SPARQL queries to be executed on the DBpedia endpoint. The main motivation is to understand user intention accurately and to map user intention and ontology conceptualization, which form the model for concepts and their interrelationships in an ontology (Xu et al. 2014).

Intui3 is proposed as a multilingual question answering platform over linked data that applies both syntactic and semantic analysis of natural language input while formulating RDF triples to represent user queries. Dima addresses the problems in traditional keyword-based searches – that results are given only as a ranked list of documents rather than a precise answer - and proposes a solution by offering a new alternative search paradigm. The Intui3 search paradigm is based on a predicate and entity index and descriptions of all entities and predicates are sourced from DBpedia (Dima 2014).

In the study of (He et al. 2014), a question answering system over linked data, namely CASIA@V2, is described. A Markov Logic Network, a joint learning model, is used to detect and classify phrases and perform mapping of phrases with semantic entities (Richardson and Domingos 2006). In the first step, the input question is decomposed to detect phrases, followed by mapping the selected candidate phrases and concepts in DBpedia. Decomposed phrases that contribute to user intention are grouped to generate triples which are actually the components of SPARQL.

(Park et al. 2015) propose ISOFT as a SPARQL template generator for natural language questions that combines two notable fields: knowledge-based question answering and information retrieval-based question answering. Semantic similarity is positioned as the focus of this study. The natural language input query is parsed into constituent phrases to be matched with SPARQL templates by imposing semantic similarity. Phrases interpreted as a contributor to answer generation are extracted to be mapped with uniform resource identifiers (URIs) that represent concepts in the ontology.

The study of (Paredes-Valverde et al. 2015) proposes an ontology-based information retrieval system called ONLI (Ontology-based Natural Language Interface). The structure and context of a natural language question are represented with an ontology model in DBpedia to address user intention. Questions are classified regarding their context. The probability of returning accurate answers is improved by classifying the NL questions to reduce search space. The ONLI pipeline has 3 main modules: question processing, knowledge bases, and answer searching and building. The question processing module deals with the pre-processing of natural language input syntactically to further improve the results of semantic analysis. Identifying semantically similar terms or finding synonyms enables semantic enrichment of the input. An ontological model, namely the Question Model, is used to organize answer searching and building. Applied search methods vary in different domains. ONLI provides possible answers with relevance ranking.

QAnswer is a question answering framework developed by (Ruseti et al. 2015) to convert natural language input questions into a SPARQL query to be executed on a DBpedia endpoint. The first task in their pipeline is to detect the DBpedia entities and the type of dependency relationships between them. The outcome of the dependency-parsing phase is a directed graph with vertices and edges, where vertices represent annotated tokens and edges hold collapsed dependencies. After addressing the dependencies, entity type detection is performed. After deciding on the various types and dependencies, candidate matches are generated. In other words, multiple different directed candidate graphs that represent possible question patterns are obtained. Finally, a scoring algorithm is applied to identify the graph with the highest score to generate SPARQL.

Baudis and Sedivy propose a question answering system pipeline called YodaQA that generates answers sourced from DBpedia and Freebase. Fundamental processes included in the pipeline can be listed as question analysis, answer production, answer analysis, answer merging and scoring, and successive refining. They formulate each question by naming three dimensions, "Clues", "Focus" and "LAT (Lexical Answer Type)". Clues represent keywords or expressions that might play essential roles in user intention. Focus is the target object to be queried. LAT is derived from focus and stands for answer type description. Semantic endpoints are searched for each clue in the sentence. Returned answers are analysed to determine LAT. Answers are classified with regard to their features. A scoring algorithm is applied for further processing with successive refining. The answer with the highest score is selected as the outcome of the pipeline (Baudiš and Šedivý 2015).

Mazzeo and Zaniolo put forward a question answering framework named CANaLI (Context-Aware controlled Natural Language Interface) to return results for controlled natural language (CNL) questions. CNL refers to restricting the grammar to make the language more easily machine-interpretable and creating a formal structure to the input questions while still protecting the natural format of the language. CANaLI suggests auto-generated question patterns to end-users typing questions, thereby correcting them semantically, syntactically, and grammatically in accordance with the concepts in the underlying knowledge base. DBpedia, MusicBrainz, DrugBank, Diseasome, and SIDER are the domain knowledge bases. Encyclopedic, music, and medicine are employed as data sources (Mazzeo and Zaniolo 2016).

AMAL (Ask Me in Any Language) is a solution that allows users to ask questions in French that are then automatically converted into SPARQL queries to find answers found in DBpedia (Radoev et al. 2017). AMAL has four main phases in generating an answer: question classification, entity extraction, property identification, and SPARQL query building. For now, it only focuses on simple questions that hold one single entity and a single property. However, the authors define AMAL as a still-evolving system that has future capacity for complex questions which hold more than one entity-property match.

(Diefenbach et al. 2018) proposes a question answering component entitled WDAqua-core0 that finds answers to natural language queries and keyword queries over DBpedia and Wikidata. DBpedia provides language support only for English, whereas Wikidata extends the range to French, German and Italian. WDAqua-core0 serves the research community through the Qanary Ecosystem (Diefenbach et al. 2017b), which is a framework to integrate question-answering components for reusability. The authors apply a combinatorial approach based on the semantic items represented in linked data sources. Two triple patterns, SELECT and ASK, are handled by WDAqua-core0. Only using the COUNT operator and not being able to handle questions containing comparative and superlative expressions are expressed as limitations of their study.

A method called SPARQLtoUser is described in another study by (Diefenbach et al. 2017a). It focuses on generating a user understandable version of SPARQL by converting user intention to a representation of a SPARQL query. The noteworthy features of SPARQLtoUser are claimed to be its multilingualism and portability. Users can generate answers to questions in different languages from different knowledge bases. Instead of verbalizing the query, a generic approach to building up a user understandable schematic representation of the query is proposed in their study.

The above-mentioned studies and this study are compared according to year, domain, language, learning appliance, and employed methods in Table 1.

Table 1. Comparison of this study (Geo-TR) and literature studies

| System | Year | Domain | Language | Learning Applied or Not | Methods |
|---|---|---|---|---|---|
| SWSNL | 2013 | Multidomain (tested on accommodation, medicine, and DBpedia) | Multilingual (tested in Czech and English) | Yes | POS Tagging, Dependency Analysis, Named Entity Recognition, Lemmatizing |
| Xser | 2014 | Multidomain (tested on Freebase and DBpedia) | Multilingual (tested in English) | Yes | POS Tagging, Dependency Analysis, Named Entity Recognition |
| Intui3 | 2014 | Multidomain (tested on DBpedia) | English | No | POS Tagging, Dependency Analysis, Named Entity Recognition, Lemmatizing, Chunking |
| CASIA@V2 | 2014 | Multidomain (tested on DBpedia) | English | Yes | POS Tagging, Dependency Analysis, Lexicon Model |
| ISOFT | 2015 | Multidomain | English | No | Natural language input query, Imposing semantic similarity, Matching with SPARQL templates |
| ONLI | 2015 | Multidomain (tested on DBpedia) | Multilingual (tested in English) | No | POS Tagging, Named Entity Recognition, Lemmatizing, Synonym Extension |
| QAnswer | 2015 | Multidomain (tested on DBpedia) | English | No | Dependency Analysis, Named Entity Recognition, Generate SPARQL |
| YodaQA | 2015 | Multidomain (tested on Freebase and DBpedia) | English | Yes | POS Tagging, Dependency Analysis, Named Entity Recognition, Lexicon Model |
| CANaLI | 2016 | Multidomain (tested on DBpedia, music, and medicine) | English | No | Controlled Natural Language Grammar, Ontology-Driven Auto-Completion |
| Amal | 2017 | Multidomain (tested on DBpedia) | French | Yes | Named Entity Recognition, Ontology-Driven Property Extraction, Lexicon Model |
| WDAqua-core0 | 2017 | Multidomain | Multilingual (tested in English, French, | No | Named Entity Recognition, |

| | | (tested on DBpedia and Wikidata) | German and Italian) | | Ontology-Driven Property Extraction |
|---|---|---|---|---|---|
| SPARQLtoUser | 2018 | Multidomain (tested in English) | Multilingual | No | The extended version of WDAqua-core0 with user interaction |
| **OUR STUDY (Geo-TR)** | 2019 | Multidomain (tested on Geographic domain) | Multilingual (tested in Turkish) | Yes | POS Tagging, Dependency Analysis, Named Entity Recognition, Ontology-Driven Property, and Instance Validation |

### 3. Ontology Development (Natural Language Aware)

The term "ontology" comes from the combination of the ancient Greek words "ontos", which means "being", and "logos", which means "word". In philosophy, ontology is defined as the subject of existence by (Gaševic et al. 2006), and existence categorization for a specific domain by (Sowa 2000). Ontologies are applied to represent knowledge by defining a set of concepts and relationships in computer science. By modelling a specific domain with concepts, including their properties, and relationships, an ontology represents a structural shared vocabulary among multiple systems.

Underpinning the development of an ontology is the requirement that it allows for the sharing of domain knowledge to provide semantic interoperability and facilitates the reusing of ontological concepts in different platforms. The most critical factor during its development is understanding the set of questions and related answers that an ontology must be able to answer correctly. These are competency questions. (Uschold and Gruninger 1996; Grüninger and Fox 1995). User stories, possible use case scenarios, and normal and alternative processing steps are helpful to determine such competency questions. Compared with other steps involved in ontology development, dealing with competency questions is essential, not only to define the context of the ontology but also to underpin the architecture behind it. Information contexts that cover competency questions provide clues about how to encode semantic items in the ontology and any application platform. Additionally, competency questions are dynamic. For long-term sustainability purposes, they must be updated by extending the context of the ontology (Kendall and McGuinness 2019). Competency questions help an ontology to be natural language-aware, which fits well with question answering systems that take input in the form of natural language.

In this study, a natural language-aware geographical ontology was developed by applying the steps defined in Ontology Development 101 (Noy and McGuinness 2001). Instead of using an existing ontology encoded in a language other than Turkish and having to deal with the burden of translation, an ontology (named GEO-TR) was developed in respect of the geography of Turkey. All data and object properties, individuals, and class names are in Turkish. Additionally, the development of a Turkish ontology in the geographical domain contributes to the field and addresses a gap in the current literature.

The first step in ontology development 101 is determining the domain and scope by defining competency questions. The questions: "Which type of domain will the ontology cover?", "For what the ontology will be used?" and "What types of questions will be answered by the ontology?" should be answered during the first step of the development. While designing the GEO-TR ontology, Chapter 6: "Spatial Synthesis of Turkey" from the 10$^{th}$ grade geography class from the national curriculum was chosen as the target scope. Competency questions were derived from "Characteristics and distribution of landforms", "Climate of Turkey", "Features of geospatial model of Turkey" and "Water resources of Turkey (rivers, seas, lakes)" which are the subsections of this chapter. Sample competency questions from these subsections can be listed as: "Lütfen, Türkiye'deki şehirleri listeler misiniz? (Please list the cities in Turkey?)", "İzmir'in komşularını gösterir misin? (Which provinces border Izmir?)", "Akdeniz bölgesinde bulunan dağları gösterir misin? (List the mountains in the Mediterranean region?)", "Manisa şehrinin çevresinde hangi şehirler konumlanır? (Which cities are located in the province of Manisa?)", "İzmir' in en yüksek dağı hangisidir? (What is the highest mountain in Izmir?)", "Ege Bölgesi'ndeki nehirlerin uzunluklarını gösterir misin? (Can you tell the length of the rivers in the Aegean region?), "Türkiye' de en fazla yağış alan il hangisidir? (Which city has the most rainfall in Turkey?)".

Step 2 was the consideration of reusing of existing ontologies. This was rejected for language-related reasons. Instead, the decision was taken to develop an ontology from scratch. Step 3 concerned the determination of competency questions. This involved determining the main conceptualization, scope, and hierarchies in the ontology to avoid overlap between concepts. Concepts, corresponding properties, and candidate relations were then defined.

Important concepts and related terms in the GEO-TR ontology are set out in Table 2.

**Table 2.** Main concepts and related terms

| Important Concepts | Related Terms |
|---|---|
| Ada (Island) | konumlanir (locatedIn), nufus (population), |
| Bogaz (Strait) | konumlanir (locatedIn), uzunluk (length) |
| Bolge (Region) | konumlanir (locatedIn), konumVar (hasLocations), nufus (population), yuzolcumu (surface area) |
| Dag (Mountain) | konumlanir (locatedIn), yukseklik (height) |
| Deniz (Sea) | konumlanir (locatedIn), derinlik (depth), tuzluluk (salinity) |
| Gol (Lake) | konumlanir (locatedIn), derinlik (depth) |
| Nehir (River) | konumlanir (locatedIn), uzunluk (length) |
| Ova (Plain) | konumlanir (locatedIn), yuzolcumu (surface area) |
| Sehir (City) | konumlanir (locatedIn), konumVar (hasLocations), nufus (population), yuzolcumu (surface area), yukseklik (height), ortalamaYagis (average rainfall), komsu (neighbourOf) |
| Ilce (District) (subclass of Sehir) | konumlanir (locatedIn), nufus (population), yuzolcumu (surface area) |
| Ulke (Country) | konumlanir (locatedIn), konumVar (hasLocations), nufus (population), yuzolcumu (surface area), iklim (climate), baskent (capital) |

After determining the main conceptualization and relationships, Step 4 involved defining classes and applying a corresponding hierarchy. Within GEO-TR, important terms are represented as classes. Each of these constitutes a subclass of the "Thing" class that represents the root node in the ontology. Ada (Island), Bogaz (Strait), Bolge (Region), Dag (Mountain), Deniz (Sea), Gol (Lake), Nehir (River), Ova (Plain), Sehir (City), Ilce (District) (and Ulke (Country) were determined as such subclasses. The only defined hierarchy in the ontology is between the subclasses Sehir (City) and Ilce (District).

The class list and hierarchy are shown in Figures 1 and 2.

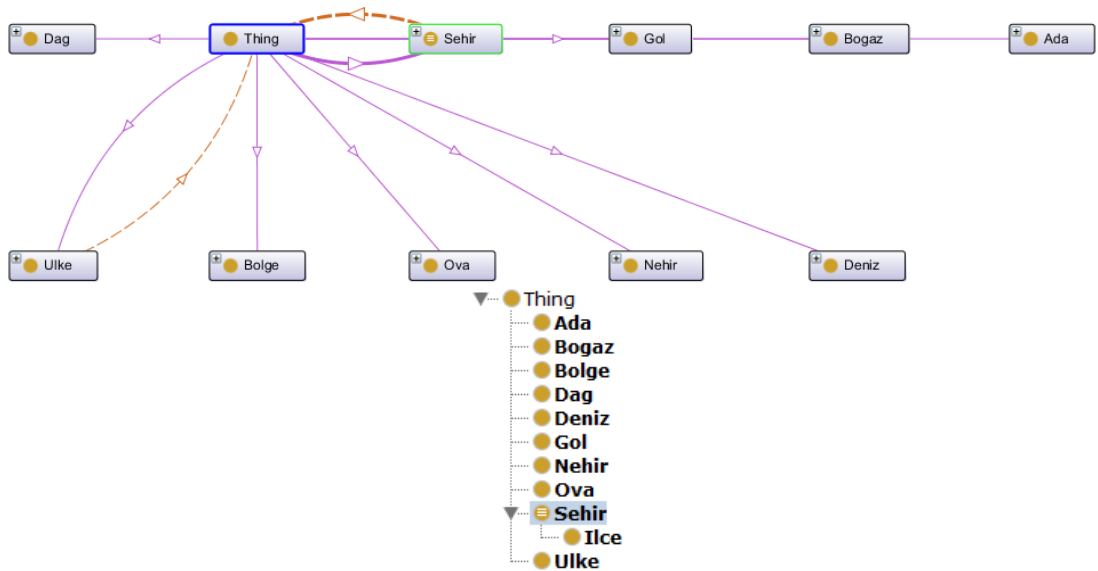

**Figure 1**. Class List in GEO-TR – OntoGraf view Protégé (Gennari et al. 2003).

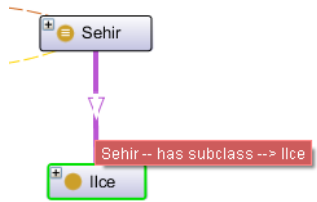

**Figure 2**. The class hierarchy between Sehir (City) and Ilce (District) – (OntoGraf view Protégé)

Step 5 of the ontology development process centred around determining the properties of classes. Related terms in the main conceptualization were selected as possible properties in the ontology. There are two types of properties in an ontology, namely data and object properties. Data properties imply a data holding element, whereas object properties hold object-oriented information. For example; a mountain has a height property. A river has a length property. The sea class's property is salinity. The city class has the properties locatedIn, neighbourOf, etc. The object and data properties of each class are illustrated in Figures 3 and 4. There is a symmetric relationship between konumlanir (locatedIn) and konumVar (hasLocations), which implies these two properties should also be defined inversely.

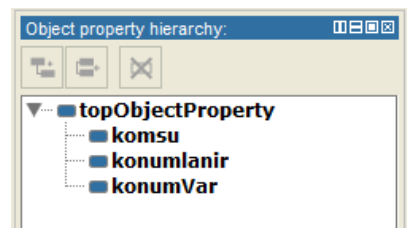

**Figure 3.** Object properties in GEO-TR

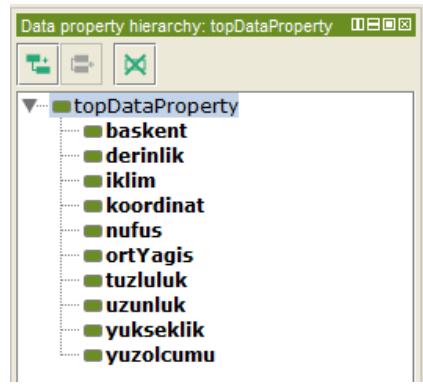

**Figure 4.** Data properties in GEO-TR

Step 6 related to defining the facets of properties that refer to value type, allowed values, cardinality (number of allowed values), and other value features a property may have. Property facets are defined as the domain and range of a property in an ontology. In ontology development terminology, the range is defined for data properties and the domain is defined for object properties. For instance, the data property nufus (population) should have an integer type that declares the range for this property. Another example, the konumlanir (locatedIn) property is valid between specific pairs like Sehir (City) – Ulke (Country) and Sehir (City) – Bolge (Region) etc. to represent the domain of this property, which defines "a city is located in a country" or "a city is located in a region". An additional example can be given for the object property komsu (neighbourOf). Classes Sehir (City), Ulke (Country), and Bolge (Region) are defined as possible domains to apply the komsu (neighbourOf) property in GEO-TR.

The final step in the process involved creating instances in the structural environment of the ontology. A sample list of instances for the classes Sehir and Bolge in GEO-TR is shown in Figures 5 and 6.

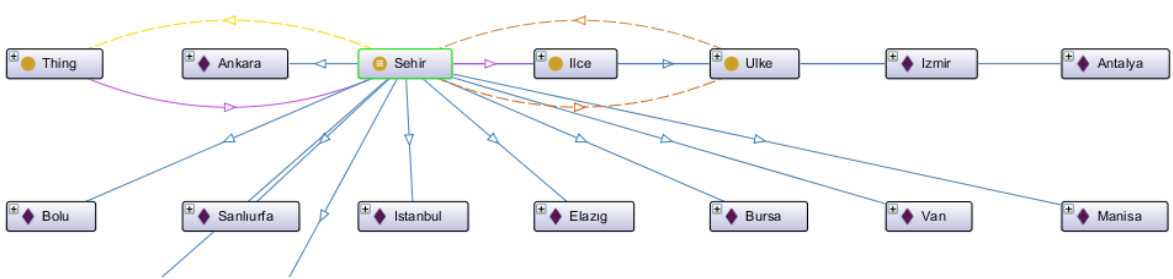

**Figure 5.** Sample list of instances of Sehir

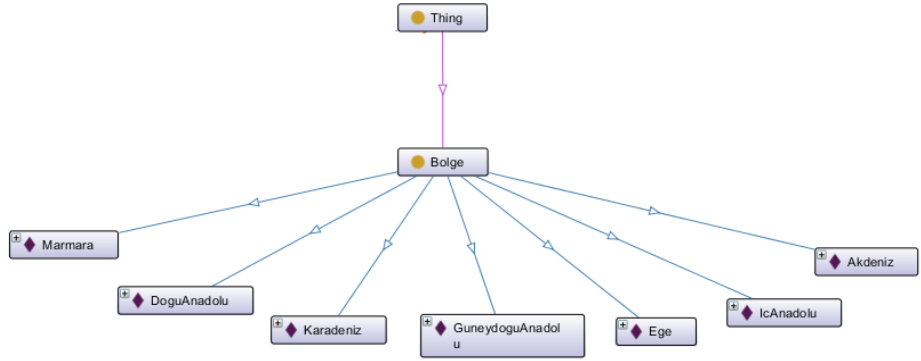

**Figure 6.** Sample list of instances of Bolge

Names of all classes, data, and object properties and instances are in Turkish. Thus, GEO-TR is coherent with semantic web-enabled geographic question-answering in Turkish, which is the principal novel contribution of this study.

## 4. Methodology

A geographical question answering framework over linked data is represented in this study for given natural language sentences in Turkish. The main components and corresponding sub-components in the system architecture are illustrated in Figure 7.

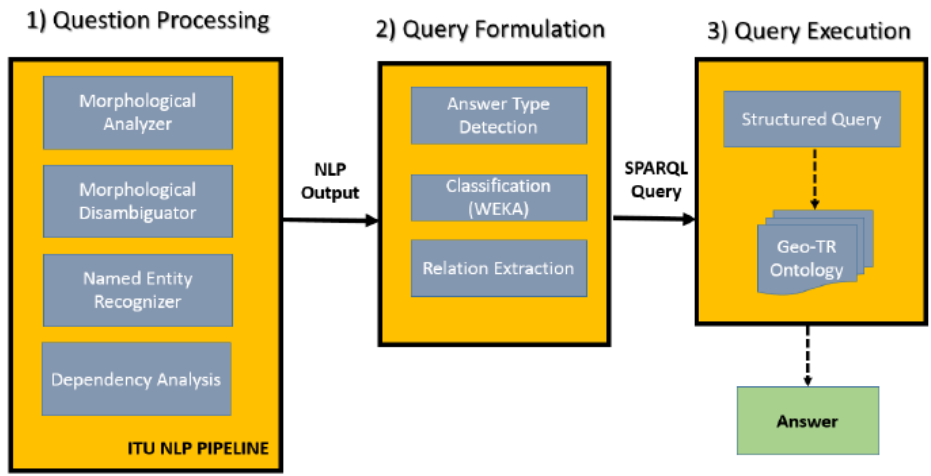

**Figure 7.** System architecture

Three main processes are configured in the system architecture, with the following layer naming: question pre-processing, query formulation, and query execution. Question pre-processing is the step in which questions are analysed morphologically and NLP techniques applied by morphologically disambiguating the POS tags of each token. In addition, named entities are recognized and dependencies between each word extracted. Each component in the question pre-processing layer acts as a pipeline, finally generating a pre-processed form of the natural language input, which is prepared to further semantically enrich the sentence. The proposed framework is designed and implemented to answer two types of questions, namely informative and quantitative reasoning involved questions. The question pre-processing layer is applied to both two types before deciding on the type of question. Next, the query formulation layer accepts the processed natural language input that is generated by the question pre-processing layer. The type of question is determined and further corresponding processing tasks are applied. A structured query in the form of SPARQL (query language of the ontology) is the outcome produced at this stage. The generated SPARQL query is executed on GEO-TR to return the answer to the query.

### 4.1 Question Pre-processing

Understanding user intention requires a combination of syntactic and semantic analysis of expressions. Eliminating tokens that do not have any contribution to achieve meaning; understanding relationships between tokens to get the focus of the sentence; tagging named entities; and extracting possible relations between these entities and the focus are the first steps in converting user intent in an unstructured form to a structured query language in the proposed study.

Turkish is a complex language that is agglutinative, morphologically different, and has free constituent order. 2 types of suffixes contribute significantly to meaning: constructive and inflectional suffixes. By adding constructive suffixes to a word, it is possible to form completely different new words semantically or words with a similar context. Inflectional suffixes are used for properly placing a word into a sentence (Erguvanli and Taylan 1984). Combinations that have a different meaning or proper usage for a given word are generated by placing the suffixes at the end of a word. The morphological features of Turkish make this language more challenging for pre-processing, necessitating a customized solution.

The Turkish NLP pipeline (Eryiğit 2014), developed and served as SaaS (software as a service) by the NLP research group of Istanbul Technical University (ITU), is used for the question pre-processing layer in Figure 7. Processing components, namely a "Morphological Analyzer", "Morphological Disambiguator", "Named Entity Recognizer" and "Dependency Parser" are utilized in this study, and spotter methods are implemented to convert the input format appropriately for each component. For a sample question input, each processing step is described in the following subsections.

### 4.1.2 Morphological Analysis

The morphological analysis step includes two main processing layers, namely determination and disambiguation of POS tags in a question input. The morphological analyzer component of the ITU Turkish NLP pipeline is the first to apply at this stage. The method, which combines the word lemmata lexicon with over 49321 entries and flag diacritics for Turkish to handle exceptions regarding phonetic and morphological rules, is presented for further processing to disambiguate the POS tags of each token. In the disambiguation layer, affixes are removed recursively without having an additional lexicon (named affix stripping) to find the accurate POS tags. Details of the method are represented in (Sahin et al. 2013). The sample question output of the morphological analysis is shown in Figure 8. Disambiguated output in Figure 8. represents the morphological structure of each word that is ready for further processing to understand their role in the sentence to achieve user intention.

### 4.1.3. Named Entity Recognition

The pipeline further processes disambiguated output with the named entity recognition method to extract location information by resolving the mentions. Several NER techniques are applied for different types of applications. The ITU Turkish NLP tool uses the methodology of the Conditional Random Fields (CRF) technique for statistical modeling (Lafferty et al. 2001) of predefined entity categories, such as person, location, organization, money, number, etc. Details of their methodology are described in their study (Şeker and Eryiğit 2012). Sample named entity recognizer output is illustrated in Figure 9.

"B-LOCATION" is the first location identifier token in which "B" stands for the beginning of the expression. 3 types of prefixes exist in NER output format. The first token of a named entity is tagged by using the prefix "B" and continues with other tokens (if possible) that are location identifiers "I-LOCATION". "I" stands for in the location expression. The last type of output prefix is "O", which represents out of any named entity tagged.

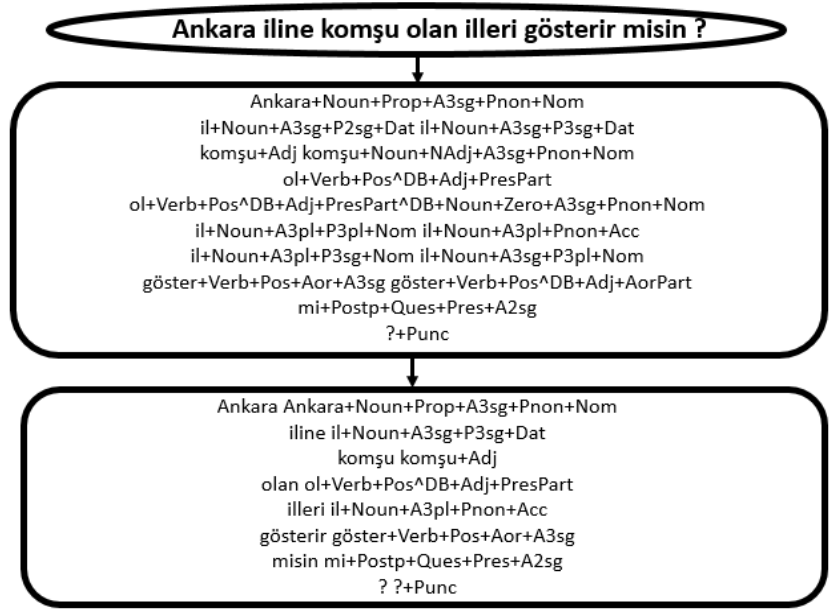

**Figure 8.** Morphological analyzer output

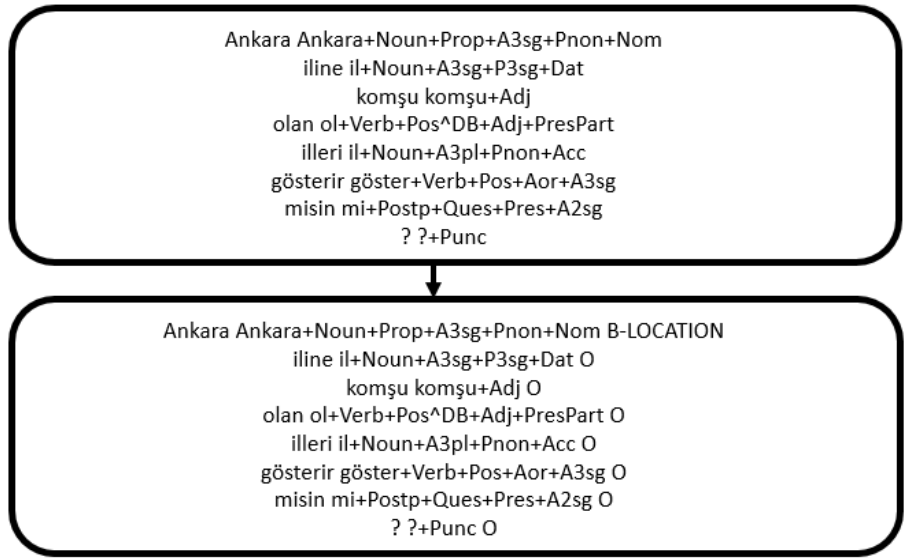

**Figure 9.** NER output

### 4.1.4 Dependency Analysis

Dependency analysis comprises tagging relationships between words to understand the roles of each token in the sentence and determining its various components, such as an object, subject, verb, or other modifiers. Generating a dependency graph composed of dependency nodes (tokens) and relationships is the underlying methodology for most dependency analysis algorithms (Nivre et al. 2010). The Conference on Computational Natural Language Learning (CoNLL-X) (Buchholz and Marsi 2006) input format and tags of the Turkish Dependency TreeBank (subject, object, modifier, classifier, possessor, etc.) are used in the dependency analysis tool of the ITU Turkish NLP pipeline (Eryiğit 2014; Eryiğit et al. 2008). The tenth CoNLL (CoNLL-X) promoted a shared training file format for multilingual dependency parsing models that has a standardized structured, column-based form. The dependency analysis result of the sample input sentence is demonstrated in Table 3.

**Table 3.** Dependency analysis output of the sample sentence

| ID | FORM | LEMMA | CPOSTAG | POSTAG | FEATS | HEAD | DEPREL | PHEAD | PDEPREL |
|---|---|---|---|---|---|---|---|---|---|
| 1 | Ankara | Ankara | Noun | Noun | Prop\|A3sg\|Pnon\|Nom | _ | 2 | _ | POSSESSOR |
| 2 | iline | il | Noun | Noun | A3sg\|P3sg\|Dat | _ | 4 | _ | MODIFIER |
| 3 | komşu | komşu | Adj | Adj | _ | _ | 4 | _ | MODIFIER |
| 4 | olan | ol | Verb | Verb | Pos^DB\|Adj\|PresPart | _ | 6 | _ | MODIFIER |
| 5 | illeri | il | Noun | Noun | A3pl\|Pnon\|Acc | _ | 6 | _ | OBJECT |
| 6 | gösterir | göster | Verb | Verb | Pos\|Aor\|A3sg | _ | 7 | _ | ARGUMENT |
| 7 | misin | mi | Postp | Postp | Ques\|Pres\|A2sg | _ | 0 | _ | PREDICATE |
| 8 | ? | ? | Punc | Punc | _ | _ | 7 | _ | PUNCTUATION |

Id is the counter to represent a token number. Form is the original token that is in the form of an original word or punctuation symbol. Lemma is the stem of a word or the same as a form if that token is a punctuation symbol. Cpostag stands for coarse-grained and Postag is the fine-grained POS tag definition from a specific treebank. Feats is the set of syntactic and morphological structure definitions, separated by "|" symbol or underscore if not available. Head and Phead values are eliminated in CoNLL-X format (Nivre et al. 2007). Deprel represents the related token and Pdeprel is the type of relationship, or in other words, type of dependency.

### 4.2 Query Formulation

The initial step of query formulation is answering type detection, entailing discovering the mention and user intention that is critically helpful in deciding answer type. Answer type classification is performed on 2 main types of questions that hold quantitative reasoning: question type 1 (QT1), or not, question type 2 (QT2). A rule-based approach detects quantitative reasoning required expressions such as "kaç tane/kaç" (how many), "ne kadar" (how many) or "en (superlative expression in Turkish - Adverb)" and further checks for the bigram of the words to detect "Adjective + Noun", "Adverb + Adjective" or "Adverb + Noun" patterns. The main flow for query formulation is shown in Figure 10.

For a given natural language input, if isQuantitative returns true, the question is determined as QT2 and query components are classified to fulfil the items in SPARQL query patterns. Target class, entity class, data property, object property, and function name are represented as categorical variables in the training model and these parameters are all matched with semantic items in GEO-TR ontology. A multilayer perceptron, which is an artificial neural network, is used to generate a learning model. The attributes and categories of the training model are shown in Table 4.

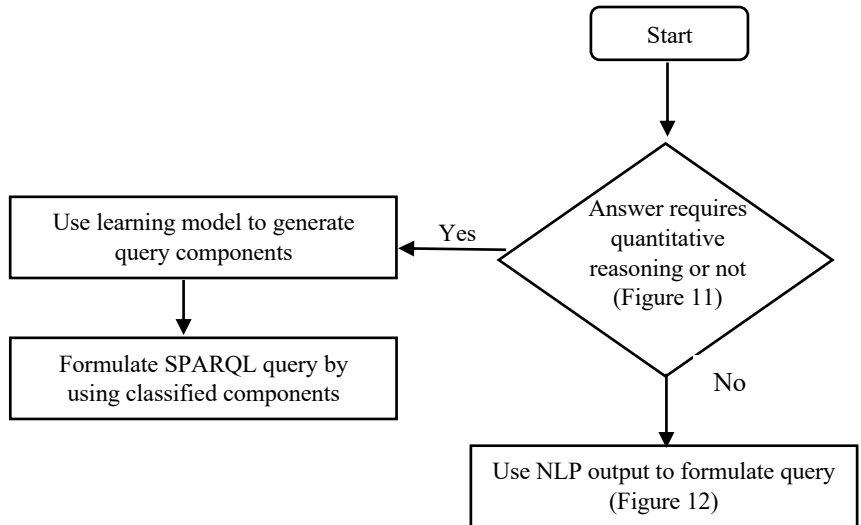

**Figure 10**. Main Flow of Query Formulation

**Table 4. Structure of training model**

| Attribute Name | Categories |
| --- | --- |
| target_class | {Sehir,Bolge,Ulke,Dag,Nehir,Gol,Ada,Ova,Deniz, Ilce,null} |
| entity_class | {Sehir,Bolge,Ulke,Dag,Nehir,Gol,Ada,Ova, Deniz, Ilce,null} |
| data_property | {yuzolcumu, populasyon, yukseklik, derinlik, tuzluluk, ortYagis, sicaklik, enlemBoylam, bitkiOrtusu, baskent, iklim, null} |
| object_property | {konumlanir,konumVar,komsu,null} |
| function_name | {count,min,max,sum,null} |

Considering the structure of the training model in Table 4, an instance sentence "Türkiye'nin en derin denizi hangisidir? (Which sea is the deepest in Turkey?)" is modelled as target_class = Deniz (Sea), entity_class = Ulke (Country), data_property = derinlik (depth), object_property = konumlanir (located in) and function_name = max (maximum as aggregate function). A SPARQL query is formulated by using classified components with corresponding query patterns. 2 types of SPAQL patterns are designed for this study. The type of aggregate function specifies the type of query pattern by using a sub query-based approach (Type 1) or not (Type 2). For the functions min and max, subquery formation is inevitable due to the nature of SPARQL queries, whereas count and sum functions do not require it (Table 5).

Table 5. Types of Query Pattern

| Query Pattern: Type 1 | Query Pattern: Type 2 |
|---|---|
| SELECT ?y ?m <br> WHERE { ?y rdf:type **ontology_name_prefix:target-class** . <br> ?y **property_prefix:data-property** ?m . <br> { SELECT (**function_name**(?var) as ?m) <br> WHERE { ?x rdf:type **ontology_name_prefix:entity-class** . <br> ?y rdf:type **ontology_name_prefix:target-class** . <br> ?y **property_prefix:object-property** ?x . <br> ?y **property_prefix:data-property** ?var <br> FILTER(regex(str(?x),"**named entity**","i")) } <br> }} | SELECT (**function_name** (?y) as ?total) <br> WHERE { ?x rdf:type **ontology_name_prefix**: **entity-class**. <br> ?y rdf:type **ontology_name_prefix**: **target-class**. <br> ?y **property_prefix**: **data-property** ?x <br> FILTER(regex(str(?x),"**named entity**","i")) } |

If any quantitative expression is not held in the natural language input, NLP output is employed in a manner based on ontology validation to formulate a SPARQL query. The algorithm deciding the answer type is indicated in Figure 11.

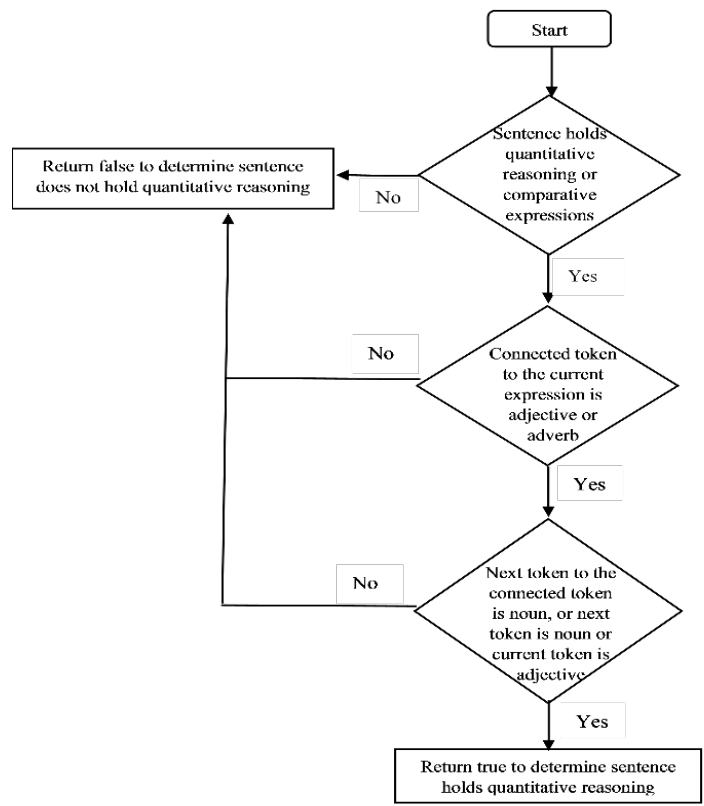

**Figure 11**. Answer Type Detection

In the case of a sentence that does not hold any quantitative reasoning expression, NLP output is combined with semantic web technologies to convert natural language input into a structured query form. The algorithm designed for the conversion is mainly based on dependency analysis,

NER output, and ontology-based validation. In the Turkish language, the target intent of the user is generally located in the object or subject of a sentence, or any other connected token with them. Following that assumption, which is based on the rules of Turkish grammar, the algorithm was designed to combine NLP output with ontology capabilities to improve accuracy while understanding user intent. The algorithm is represented and described with examples below.

**Algorithm** Algorithm to find the answer type of question in Turkish and generate SPARQL query by using processed output by NLP techniques (Method name: generateSparql)
**Require:** sentence processed by NLP techniques
*agenda:* generate query by using NLP output
**Ensure:** final_query
1: nerEntities = pipeline.getNamedEntities(nerOutput)
2: **if** *dependencyOutput.contains("OBJECT")*
3:     answerType = objectTerm
4:     axiomType = pipeline.checkAxiomType(answerType)
5:     **if** *axiomType == "CLASS"* **then**
6:         properties = pipeline.findProperties(answerType, nerEntities)
7:         *final_query* = pipeline.formulate_query(properties, nerEntities, answerType)
8:     **end if**
9:     **if** *axiomType == "DATA PROPERTY"* **then**
10:        relatedToken = pipeline.findRelatedToken(answerType, dependencyOutput)
11:        axiomTypeRelated = pipeline.checkAxiomType(relatedToken)
12:        Go back to Step 5 call the method for the input axiomTypeRelated and continue again
13:    **end if**
14:    **if** *axiomType == "OBJECT PROPERTY"* **then**
15:        relatedClass = pipeline.findRelatedToken(answerType,dependencyOutput)
16:        *final_query* = pipeline.formulate_query(answerType, nerEntities, relatedClass)
17:    **end if**
19: **if** *dependencyOutput.contains("SUBJECT")* **then**
20:    answerType = subjectTerm
21:    axiomType = pipeline.checkAxiomType(answerType)
22:    **if** *axiomType == "CLASS"* **then**
23:        properties = pipeline.findProperties(answerType, nerEntities)
24:        **if** *properties == NONE* **then**
25:            commonConnected = pipeline.findCommonConnected (dependencyOutput, answerType)
26:            Go to Step 21 call the method for the input commonConnected and continue again
27:        **end if**
28:    **end if**
29:    **else**
30:        *final_query* = pipeline.formulate_query(properties, nerEntities, answerType)
31:    **if** *axiomType == "DATA PROPERTY"* **then**
32:        relatedToken = pipeline.findRelatedToken(answerType, dependencyOutput)
33:        axiomTypeRelated = pipeline.checkAxiomType(relatedToken)
34:        Go back to Step 21 call the method for the input axiomTypeRelated and continue again
35:    **end if**
36:     **if** *axiomType == "INDIVIDUAL"* **then**

| 37: |  | connectedToken = pipeline.findConnectedToken (answerType, dependencyOutput) |
|---|---|---|
| 38: |  | axiomTypeConnected = pipeline. checkAxiomType(connectedToken) |
| 39: |  | Go back to Step 21 call the method for the input axiomTypeConnected and cont. again |
| 40: | **end if** |  |
| 41: | **if** *axiomType == "OBJECT PROPERTY"* **then** |  |
| 42: |  | commonConnected = pipeline.findCommonConnected(dependencyOutput, answerType) |
| 43: |  | Go back to Step 21 call checkAxiomType for the input commonConnected and continue |
| 44: | **end if** |  |
| 45: | **end if** |  |

The first processing step for the sample input question after applying NLP methods is deciding on the type of dependency relationship the sentence holds. As illustrated in the dependency analysis output of the sample sentence (Figure **Table 3)**, token 5 ("il" (city)) is the object of that sentence, and the axiom type of the object is checked to decide whether it is a class, a data or object property, or an individual. City is a semantic item and represented as a class in the ontology and determined as a target class for query formulation. For the condition that the object of the sentence is a class (Algorithm – Step 5), the properties of that class with the named entity (if it exists) in the sentence are found to generate a SPARQL query. "Ankara" is the named entity for this sentence (**Figure 9**). The entity class for the individual "Ankara" is extracted from the ontology as Sehir (entity class), which is the same class with the object token. Possible relationships with "Ankara" and class Sehir are extracted from the ontology. The only relationship is found to be an object property "komsu" (neighbourOf) for the sample case. The generic query pattern for the SPARQL formulation is:

SELECT ?y
WHERE { ?x rdf:type **ontology_name_prefix**: **entity-class**.
?y rdf:type **ontology_name_prefix**: **target-class**.
?y **property_prefix**: **nameOfproperty** ?x
FILTER(regex(str(?x),"**named entity**","i")) }.

By using this pattern, the query is generated as follows:

SELECT ?y
WHERE { ?x rdf:type **geo_turkce**:**Sehir** .
?y rdf:type **geo_turkce**:**Sehir** .
?y **ins**:**komsu** ?x .
FILTER(regex(str(?x),"**Ankara**","i")) }.

Another sample question includes a subject phrase via a deep-thinking algorithm. For the question "Ege Bölgesi'nin yüzölçümü ne kadardır? (How much is the total area of the Aegean region?)", the dependency analysis output is given in Table 6 and the NER result is:

**Ege ege+Noun+A3sg+Pnon+Nom B-LOCATION**

**Bölgesi'nin bölge+Noun+A3sg+P3sg+Gen I-LOCATION**

yüzölçümü yüzölçüm+Noun+A3sg+P3sg+Nom O

ne ne+Pron+Ques+A3sg+Pnon+Nom O

kadardır kadar+Postp+PCNom^DB+Noun+Zero+A3sg+Pnon+Nom^DB+Verb+Zero+Pres+A3sg+Cop O

? ?+Punc O.

Table 6. Dependency Analysis Result of Sample Sentence

| ID | FORM | LEMMA | CPOSTAG | POSTAG | FEATS | DEPREL | PDEPREL |
|---|---|---|---|---|---|---|---|
| 1 | Ege | ege | Noun | Noun | A3sg\|Pnon\|Nom | 2 | POSSESSOR |
| 2 | Bölgesi'nin | bölge | Noun | Noun | A3sg\|P3sg\|Gen | 3 | POSSESSOR |
| 3 | yüzölçümü | yüzölçüm | Noun | Noun | A3sg\|P3sg\|Nom | 5 | SUBJECT |
| 4 | ne | ne | Pron | Pron | Ques\|A3sg\|Pnon\|Nom | 5 | ARGUMENT |
| 5 | kadardır | kadar | Postp | Postp | PCNom^DB\|Noun\|Zero\|A3sg\|Pnon\|Nom^DB\|Verb\|Zero\|Pres\|A3sg\|Cop | 0 | PREDICATE |
| 6 | ? | ? | Punc | Punc | _ | 5 | PUNCTUATION |

Token 3 is the subject of the sentence and axiom type determined as a data property from GEO-TR. This is an indicator that quantitative analysis is required to handle user intent. Possible classes that are assigned with the aforementioned data property are detected in the sentence (Algorithm – Step 31). From the NER result, "Ege Bölgesi" (Aegean Region) is the named entity, and the entity class is extracted as "Bölge (Region)" from the ontology. Additionally, the dependency parsing result shows that token 2 is directly related to the subject token, and the axiom type of token 2 is a class in GEO-TR. At that point, the algorithm moves back to Step 22 to check for possible properties from the ontology but, for that sample case, the answer type is a property so searching for the commonly connected token with the subject expression is the second thing to do (Step 25). The commonly connected token is already found because of the fact that there is no other entity for that case and the algorithm formulates the query as:

SELECT ?variable

WHERE { ?x rdf:type **geo_turkce**:**Bolge** .

?x **ins**:**yuzolcumu** ?variable .

FILTER(regex(str(?x),"**Ege**","i")) }

Final sample input: "Ege Bölgesi'ndeki şehirlerin nüfuslarını gösterir misin ? (Can you show me the populations of the cities in the Aegean Region?)", is more complex and holds a determinative group expression for possessive construction ("zincirleme isim tamlaması"). The dependency analysis output is shown in Table 7 and the NER result of the sentence is as follows:

**Ege ege+Noun+A3sg+Pnon+Nom B-LOCATION**

**Bölgesi'ndeki bölge+Noun+A3sg+P3sg+Loc^DB+Adj+Rel I-LOCATION**

şehirlerin şehir+Noun+A3pl+Pnon+Gen O

nüfuslarını nüfus+Noun+A3pl+P3sg+Acc O

gösterir göster+Verb+Pos+Aor+A3sg O

misin mi+Postp+Ques+Pres+A2sg O

? ?+Punc O

**Table 7.** Dependency Analysis Output of Final Sample Sentence

| ID | FORM | LEMMA | CPOSTAG | POSTAG | FEATS | DEPREL | PDEPREL |
|---|---|---|---|---|---|---|---|
| 1 | Ege | ege | Noun | Noun | A3sg\|Pnon\|Nom | 2 | POSSESSOR |
| 2 | Bölgesi'ndeki | bölge | Noun | Noun | A3sg\|P3sg\|Loc^DB\|Adj\|Rel | 5 | MODIFIER |
| 3 | şehirlerin | şehir | Noun | Noun | A3pl\|Pnon\|Gen | 4 | POSSESSOR |
| 4 | nüfuslarını | nüfus | Noun | Noun | A3pl\|P3sg\|Acc | 5 | OBJECT |
| 5 | gösterir | göster | Verb | Verb | Pos\|Aor\|A3sg | 6 | ARGUMENT |
| 6 | misin | mi | Postp | Postp | Ques\|Pres\|A2sg | 0 | PREDICATE |
| 7 | ? | ? | Punc | Punc | _ | 6 | PUNCTUATION |

After determining token 4 ("nüfuslarını" (population)) as the object of the sample sentence, the algorithm firstly checks the axiom type for the stemmed form of the token or synonyms (nüfus (population)). The axiom type is determined as a data property and the algorithm continues with Step 10 by using the findRelatedToken method to find the directly dependent token with the token 3 object ("şehirlerin"). The dependency output is critical for specifically this type of question in order to understand user intent to find the population of cities located in the Aegean region rather than displaying the population of the Aegean region. Token 3 "şehir" (city) is checked for the axiom type from the ontology and the result returns as the class that moves the algorithm to Step 6 by assigning the answerType to the related token ("şehir"). Properties defined between "Ege Bölgesi" and "şehir" are extracted from the ontology and only one object property returns, namely "konumVar (hasLocation)". The named entity, entity class, target class, data, and object properties are assigned to formulate the query as follows:

SELECT ?variable

WHERE { ?x rdf:type **geo_turkce**:**Sehir** .

?y rdf:type **geo_turkce**:Bolge .

?y **ins**:**konumVar** ?x .

?x **ins**:**populasyon** ?variable .

FILTER(regex(str(?y),"**Ege**","i"))

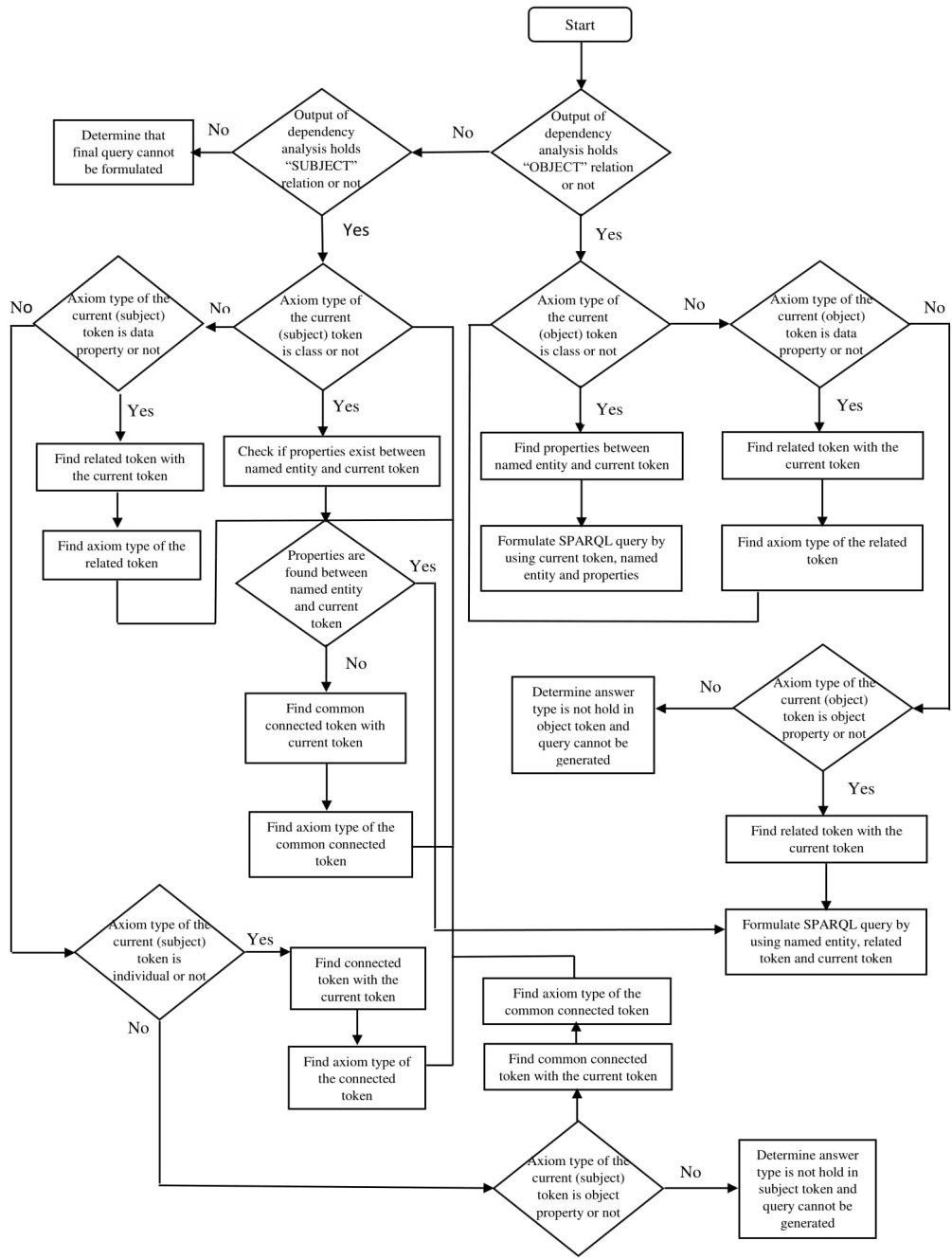

**Figure 12**. Flowchart of Algorithm

### 4.3 Experimental study

The experimental study was performed on two types of questions (QT1 and QT2: See Section 4.2). A comparison was generated to give the results for QT1 in order to understand more deeply the contribution of the NLP output while interpreting the user input. Our main contribution is to show the results of combining the NLP output with semantic technologies so as to build up a SPARQL query by discovering accurate entities and relationships. Matching the entities and relationships in the NLP and ontology by double-checking the tokens, and dependencies between them, is the main focus of the study. Instead of only checking for each token, their types, and possible relationships between them in the ontology, the hybrid method disambiguates the possible relationships by applying NLP output. A basic user interface, as illustrated in Figure 13, was implemented for an experimental study of 100 test questions.

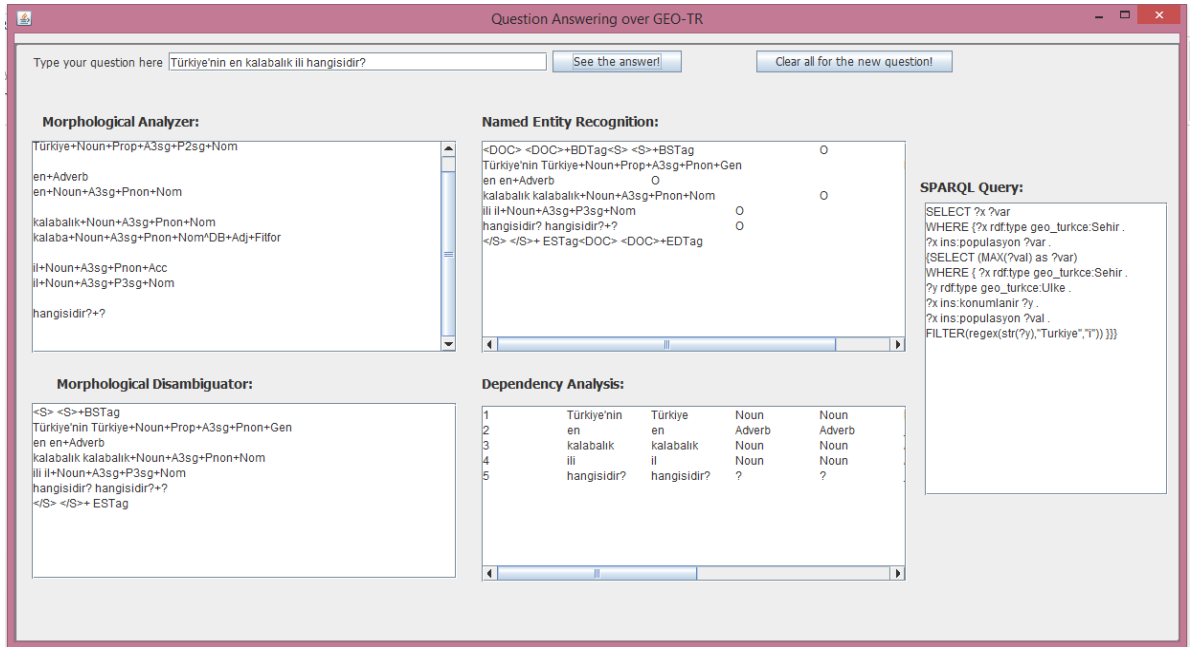

Figure 13. User Interface for Experimental Study

### 4.4 Results and Discussion

Comparison paradigms consist of using a hybrid approach (NLP + ontology-based approach) and only applying an ontology-based approach. Comparison metrics given in Table 8 are precision, recall, and F-measure.

Table 8. Results of Comparison

| Method | Precision | Recall | F-Measure |
|---|---|---|---|
| **Method 1:** Hybrid approach | 0.77 | 0.68 | 0.71 |
| **Method 2:** Ontology based approach | 0.64 | 0.57 | 0.60 |

Results indicate that using morphologically analyzed word forms and their dependencies with semantic items in GEO-TR contributes to improving the accuracy of the framework. Questions with possessive constructions are appropriate examples of how NLP output directly contributes to the meaning by applying dependency analysis to decide on target answer type. Relationships between words that might be critical to ascertaining the answer type are missed by Method 2. For instance, for a question such as "Ege Bölgesi'ndeki şehirlerin nüfuslarını gösterir misin? (Can you show the populations of cities in Aegean Region?)", which has a possessive construction, the ontology-based approach failed to disambiguate whether the population of the Aegean Region (individual in GEO-TR) was intended, or populations of cities that are located in the Aegean Region separately. The relationship "POSSESSOR" between the tokens "şehirlerin(cities)" and "nüfuslarını (populations)" disambiguates the user intent as the population of the cities are asked for. The only advantage of Method 2 relates to capturing entities, labels that are not recognized by the named entity recognition process. Considering overall performance, this advantage is not sufficient to compete with the hybrid method. The ontology-based method simply checks for each, whether it exists in the ontology or not. If a token is found in the ontology, possible relationships and types of axiom are checked. By using the same SPARQL patterns with Method 1, fulfilling the items with corresponding tokens is performed.

Another drawback of Method 2 is that it is challenging to decide on the answer type for sentences that involve more than one class item. A good example can be given as: "İzmir şehri hangi bölgededir?". "Izmir" (individual), "Sehir" (class), and "Bolge" (class) are matched with semantic items in GEO-TR after processing. Possible extracted relationships between individuals and classes are "konumlanir (locatedIn)" (Izmir – Bolge) and "komsu (neighbourOf)" (Izmir – Sehir). An incorrect intention can be produced by using the outcome of Method 2 to show the neighboring cities of Izmir. Overall, the application of Method 1 results in more accurate results for all facts compared to Method 2.

A learning model generated by the supervised learning method can be used to predict query components. Predicted components are target class, entity class, data property, object property, and aggregate function name to perform quantitative reasoning analysis for the given attributes. Predicted components are used to formulate the SPARQL query. The accuracy of the learning model that is built to handle QT2 is 0.72. The experiment is performed by using a train/test split of 0.8/0.2.

## 5. Conclusion

In order to fill a gap in the literature, a Turkish question answering framework over linked data (GEO-TR) in the geographic domain is proposed in this study. Combining NLP techniques and an ontology, two types of questions (QT1 and QT2) are handled in this framework. The main conclusion is that a hybrid approach (Method 1) interprets a sentence in natural language more accurately than an ontology-based approach (Method 2). Another significant contribution of this study is the creation of a novel Turkish ontology in the geographical domain, which has been developed by following the rules of ontology development 101(Noy and McGuinness 2001).

GEO-TR is extendable and ready to use as a data source for other researchers. Following the question pre-processing and query formulation, the experimental study demonstrates the main contribution of hybrid architecture, and results are given by using precision, recall, and F-measure values.

Currently, this study is not capable of handling more complex queries with more than one recognized entity, more than one level possessive constructions, or conditional and comparative expressions. Future work is suggested to apply deep learning techniques to handle complexity in question forms. Moreover, extending the coverage and creating a multi-ontology platform could be an additional direction of future study. It is proposed that a pipeline that accepts natural language input and classifies the sentence according to the domain types and matches with a corresponding ontology can fit with this architecture.

**Declarations:**
**Funding** This research received no external funding.
**Conflicts of interest/Competing interests** The authors declare no conflict of interest, financial or otherwise.
**Availability of data and material** Not applicable
**Code availability** Not applicable
**Acknowledgments** Declared none.